\documentclass[letterpaper]{article} 
\usepackage{aaai25}  
\usepackage{times}  
\usepackage{helvet}  
\usepackage{courier}  
\usepackage[hyphens]{url}  
\usepackage{graphicx} 
\usepackage{natbib}  
\usepackage{caption} 

\usepackage{algorithm}

\usepackage{newfloat}
\usepackage{listings}
\DeclareCaptionStyle{ruled}{labelfont=normalfont,labelsep=colon,strut=off} 
\floatstyle{ruled}
\newfloat{listing}{tb}{lst}{}
\floatname{listing}{Listing}
\floatstyle{ruled}
\newfloat{listing}{tb}{lst}{}
\floatname{listing}{Listing}
\usepackage{inconsolata}
\usepackage{times}
\usepackage{url}
\usepackage{latexsym}
\usepackage{booktabs}
\usepackage{algorithm}
\usepackage{algpseudocode}
\usepackage{tikz}
\usepackage{todonotes}
\usepackage{booktabs}
\usepackage{multirow}
\usepackage{booktabs}
\usepackage{amssymb}
\usepackage{subcaption}
\usepackage{adjustbox}
\usepackage{hhline}
\usepackage{multirow}
\usepackage{hhline}
\usepackage{longtable}
\usepackage{hhline}
\usepackage{amsmath}
\usepackage{pifont}
\newcommand{\xmark}{\ding{55}}%
\definecolor{mygreen}{rgb}{0.0, 0.5, 0.0}
\definecolor{myred}{rgb}{0.64, 0.0, 0.0}
\newcommand\blfootnote[1]{
    \begingroup
    \renewcommand\thefootnote{}\footnote{#1}
    \addtocounter{footnote}{-1}
    \endgroup
}
\nocopyright 

\title{Confidence-Aware Deep Learning for Load Plan Adjustments in the Parcel Service Industry}

\author{
    Thomas Bruys \quad Reza Zandehshahvar \quad Amira Hijazi \quad Pascal Van Hentenryck \\
    \normalfont NSF Artificial Intelligence Research Institute for Advances in Optimization, Atlanta, USA \\
    Georgia Institute of Technology, Atlanta, USA \\
    \texttt{\{thomas.bruys, ahijazi6\}@gatech.edu \\ \{reza, pascal.vanhentenryck\}@isye.gatech.edu}
}

\date{}

\begin{document}
\maketitle
\begin{abstract}
    This study develops a deep learning-based approach to automate inbound load plan adjustments for a large transportation and logistics company. It
    addresses a critical challenge for the efficient and resilient
    planning of E-commerce operations in presence of increasing
    uncertainties. The paper
    introduces an innovative data-driven approach to inbound load
    planning. Leveraging extensive historical data, the paper presents a
    two-stage decision-making process using deep learning and conformal
    prediction to provide scalable, accurate, and confidence-aware
    solutions. The first stage of the prediction is dedicated to tactical
    load-planning, while the second stage is dedicated to the operational
    planning, incorporating the latest available data to refine the
    decisions at the finest granularity. Extensive experiments compare
    traditional machine learning models and deep learning methods. They
    highlight the importance and effectiveness of the embedding layers for
    enhancing the performance of deep learning models. Furthermore, the
    results emphasize the efficacy of conformal prediction to provide
    confidence-aware prediction sets. The findings suggest that data-driven methods can
    substantially improve decision making in inbound load planning,
    offering planners a comprehensive, trustworthy, and real-time
    framework to make decisions. The initial deployment in the industry
    setting indicates a high accuracy of the proposed framework.
\end{abstract}

\blfootnote{This research was supported by NSF award 2112533}

\section{Introduction}
In recent decades, the trucking industry has experienced significant
changes driven by the rapid growth of E-commerce, driven by increased
online shopping and reduced reliance on physical stores, thereby
altering logistics and delivery demands. According to the American
Trucking Association, in 2022, 72\% of total US domestic tonnage was
shipped via trucks, which corresponds to 11.46 billion tons of freight
\citep{trucktonnage}. This represents a significant increase over the
past ten years, up from 8.19 billion tons in 2012
\citep{trucktonnage2012}. This dramatic surge in demand has profoundly
impacted parcel service industries, creating an urgent need for
advanced logistics capabilities to efficiently plan and manage
shipments across large-scale logistics networks, while meeting
customer expectations.

Load planning is crucial to the operations of these logistic networks,
as it defines the transportation capacity of the network 
\citep{Bakir2021}. At the tactical level, e.g., a week before the day
of operations, load planning determines the number of trailers to
allocate across the network to serve the forecasted
demand. Tactical planning typically uses average estimates for
demand. However, actual demand can vary significantly, rendering the
tactical plan less effective on the day of operations
~\citep{lindsey2016improved}. To address this challenge, parcel
service companies adjust their tactical plans on the day of operations
by re-routing shipments on alternative paths, modifying the arrivals
and due times of some shipments, or shifting inbound
loads~\citep{crainic2000service}.

This paper is motivated by the need of an industry partner, one of the
largest parcel shipment companies, for a decision-support tool to
automate inbound load-shifting process. Load shifting is the process
of re-directing inbound loads to a different destination. Currently, these decisions are being made manually,
requiring planners to evaluate both load-specific and
destination-specific data to make quick decisions about where to
process each load.  To alleviate the burden on human planners, this
paper proposes a machine learning (ML) decision-support tool to
predict the processing building and sort of each planned load (i.e.,
the location and time that the loads are planned to be processed). The proposed model acts as a proxy for the automation of load-shifting decision-making which enables faster, more
consistent, and reliable decisions, while reducing the burden on human
planners. Moreover, the machine learning tool can learn from both
historical decisions made by human planners and optimal decisions
generated by optimization programs, allowing for both high-quality
decisions and real-time decision support.

The key contribution of the paper is a two-stage decision-support
framework, that captures the needs of both tactical and operational
load planning. At the tactical level (stage 1), one week prior to the
day of operation, the model predicts the building and sort for each
load. This allows planners to anticipate potential load shifts in
advance and thus enables efficient resource allocation. At the
operational level (stage 2), the model provides real-time adaptability
on the day of operation by refining sort predictions based on updated
forecasts, such as the estimated arrival time of each load. Attempting
to refine the building prediction on the day of operations is not
feasible, as it would create delays and violate service
commitments. {\em The proposed two-stage framework effectively balances the
need for proactive planning with the flexibility to adapt to real-time
conditions.}

A key aspect of the approach is to integrate training data from
buildings across multiple clusters, enriching the decision-support
tool with a network-wide perspective based on the features of the
planned destinations. This ensures that {\em decisions at the building
level are informed by a comprehensive understanding of the entire
network}, leading to more effective load planning.

The framework also incorporates conformal prediction to provide
confidence-aware recommendations. Relying on single-point predictions
can lead to disruptions in the downstream logistics process if these
predictions are incorrect, reducing the trustworthiness of the ML
model. Conformal prediction addresses this issue by providing a
prediction set with theoretical guarantees, ensuring that the true
processing location and sort are included within the recommendations
with a specified confidence level.

\subsection{Related Work}
\subsubsection{Load Planning\\}

This paper contributes to the literature on service network design for
freight transportation, specifically focusing on load planning. While
load planning at the tactical level has been widely studied, there has
been less emphasis on operational planning.

Tactical load planning involves determining the number of trailers to
dispatch between each pair of terminals as well as the time of
dispatch. \cite{continuous} modeled this problem as a space-time
network and proposed an iterative refinement algorithm based on
dynamic discretization discovery (DDD). \cite{enhanced} enhanced DDD
by using acceleration techniques, including relaxations, valid
inequalities, and symmetry breaking. For comprehensive survey on
tactical load planning, the readers are referred to
\cite{CRAINIC2000272}, \cite{Wieberneit2007ServiceND}, and
\cite{current_research}.

Operational planning involves the adjustment of the tactical plan by
re-routing shipments on alternative paths, modifying the arrivals and
due times of some shipments, or shifting inbound loads to hedge
against day-to-day demand changes and operational constraints
\citep{crainic2000service}. The value of re-routing shipments on $p$
alternative paths is highlighted in \cite{inproceedings}. Building on
this, \cite{HERSZTERG20221021} proposed a near real-time solution for
operational planning that effectively considers one alternative
path. More recently,
\cite{ojha2024optimizationbasedlearningdynamicload} developed an ML
proxy for dynamic outbound load plans, enabling the operational load
plan to be adjusted dynamically as package volumes change.

The aforementioned studies primarily addressed outbound
decision-making. On the inbound side, \cite{MCWILLIAMS2009958}
introduced a framework for optimizing unload schedules of the inbound
loads to minimize the time-span of transfer operations using a dynamic
load-balancing algorithm. Additionally, \cite{bugow} employed a
genetic algorithm to optimize inbound truck scheduling.

\subsubsection{ML in Parcel Service Industry\\} 
In recent years, the increasing
availability of data has driven the popularity of applying ML methods
across various domains. To date, ML has primarily been applied in the
parcel service industry for arrival time estimation, demand
forecasting, vehicle routing, and containerization, as discussed in
the survey by \cite{TSOLAKI2023284}. For example, \cite{Munkhdalai}
presented a Deep Learning (DL) approach for demand forecasting in the
parcel service industry during special holidays, while \cite{ye}
proposed a graph-learning based method coupled with a temporal
attention module for inbound parcel volume
forecasting. \cite{ESKANDARZADEH2024103519} demonstrate a hybrid ML
and optimization approach for parcel sortation and containerisation
decisions.

Although ML methods hold considerable potential, their practical
deployment often faces challenges, such as insufficient trust and
inconsistent reliability. ML models can sometimes
make incorrect decisions with high confidence, compounding these
concerns. This paper addresses these issues by leveraging conformal
prediction, which is a distribution-free and model-agnostic approach
\citep{vovk, romano2020classificationvalidadaptivecoverage,
  angelopoulos2022uncertaintysetsimageclassifiers}. Within the context
of parcel service industry, conformal prediction has been utilized for
uncertainty quantification of the time and location of pick-up
requests made by customers to couriers \citep{yanetal}. However, this
is still an under-explored area of research in the parcel service
industry.

\subsection{Contributions}

To the best of our knowledge, this paper is the first to introduce a
confidence-aware data-driven framework to facilitate fast, reliable,
and scalable inbound decision making for both tactical and operational
load planning. More precisely, for each load, the proposed approach
predicts the inbound destination of each load by utilizing a two-stage
decision-making framework. The first stage takes place one week before
the day of operations and predicts the building and sort for
processing each load. This tactical-level prediction is crucial as it
allows planners to identify loads that must be shifted to different
buildings well in advance, enabling the proper resource
allocation. The second stage takes place on the day of operations and
refines the sort prediction based on updated forecasts, including the
estimated arrival time of each load. The two-stage approach is
motivated by the lack of reliable data on the planned arrival time of
the loads a week before operations, which limits the performance of
the single-stage approach in predicting sorts.

The main contributions of the paper can be summarized as follows.

\begin{enumerate}
    \item The paper develops a deep-learning method tailored for
      tabular data to automate load plan adjustments in parcel service
      industry. The method obtains reliable predictions through a
      two-stage decision-making process, which captures both planning
      and operational constraints and objectives.

    \item The proposed framework returns confidence-aware solutions by
      enhancing the two-stage method with conformal prediction. This
      allows planners to focus on the less confident predictions,
      enhancing the overall robustness of the decision-making process
      and reducing their cognitive burden. 

     \item The paper validates the proposed framework using an
       industrial dataset collected over 1.5 years and consisting of
       more than 250,000 loads. The results demonstrate that buildings
       and sorts can be predicted with an overall accuracy of 99\% and
       87\%, respectively, one week before the day of operations. The
       second stage, with the refinement of the sort prediction on the
       day of operations, allows an improvement of 5\% in overall
       accuracy, and 20\% on the shifted loads, compared to forecasts
       made a week prior. Additionally, the conformal prediction
       results demonstrate that the proposed method provides reliable
       and typically small prediction sets, while reaching the target
       coverage rate.
\end{enumerate}

\renewcommand{\arraystretch}{1.3} 

\begin{table*}[!ht]
\centering
\caption{Nomenclature}
\label{Nomenclature}
\begin{tabular}{lp{12cm}}
\toprule
\textbf{Sets} & \textbf{Description} \\
\midrule

$\mathcal{L}$ & The set of loads (i.e., aggregation of shipments that share the same origin, destination and due date). \\
$\mathcal{B}$ & The set of buildings. \\
$\mathcal{S}$ & The set of sorts, defined based on the processing times of the loads (e.g., S1, S2, and S3.) \\
$\mathcal{G}$ & The set of building clusters (i.e., aggregation of buildings within the same state). \\

\midrule
\textbf{Parameters} & \textbf{Description} \\
\midrule

$l \in \mathcal{L}$ & Load identifier. \\
$b_{l}^{\prime} \in \mathcal{B}$ & Planned processing building of load $l$. \\
$b_{l} \in \mathcal{B}$ & Actual processing building of load $l$. \\
$s_{l}^{\prime} \in \mathcal{S}$ & Planned processing sort of load $l$. \\
$s_{l} \in \mathcal{S}$ & Actual processing sort of load $l$. \\

\midrule
\textbf{ML Nomenclature} & \textbf{Description} \\
\midrule

$\mathbf{X}$ & Input data corresponding to a set of features. \\
$\mathcal{D}_{train}$ & Training Set. \\
$\mathcal{D}_{validation}$ & Validation Set. \\
$\mathcal{D}_{calibration}$ & Calibration Set. \\
$\mathcal{D}_{testing}$ & Testing Set. \\
$y_{b}$ & Target processing building. \\
$y_{s}$ & Target processing sort. \\
$\hat{y}_{b, w}$ & Predicted processing building a week prior the day of operations. \\
$\hat{y}_{s, w}$ & Predicted processing sort a week prior the day of operations. \\
$\hat{y}_{s, d}$ & Predicted processing sort on the day of operations. \\
$\hat{y}^{cp}_{b, w}$ & Conformalized prediction set of buildings a week prior the day of operations. \\
$\hat{y}^{cp}_{s, w}$ & Conformalized prediction set of sorts a week prior the day of operations. \\
$\hat{y}^{cp}_{s, d}$ & Conformalized prediction set of sorts on the day of operations. \\
\bottomrule
\end{tabular}
\end{table*}

\renewcommand{\arraystretch}{1}

\subsection{Structure of the Paper}

The remainder of the paper is organized as follows. Section
\ref{pbm_statement} presents the problem of inbound load planning
considered in this paper. Section \ref{Data Overview} reviews the data
provided by the industrial partner. Section \ref{sec: Method} presents
the architecture of the proposed DL-based and confidence-aware
framework. Section \ref{Experimental Details} describes the case study
and experiments. The results and comparisons of different methods
including traditional ML models and DL architectures are discussed in
Section \ref{sec : Results} followed by the conclusions in Section
\ref{sec : Conclusion}.

\section{Problem Statement}\label{pbm_statement}

Table \ref{Nomenclature} presents the nomenclature and defines all
symbols and notations used throughout the paper. Logistic networks are
organized into clusters of \textit{buildings}, defined either
geographically or based on the frequency of interactions between
buildings. Within these networks, outbound trailers, also referred to
as \textit{loads}, from upstream buildings are planned to arrive and
be processed at downstream buildings during specific time windows,
known as \textit{sorts}, which divide a typical operational day into
three periods: S1, S2, and S3.

Consider a set of inbound loads \(\mathcal{L}\) that are planned to be
sorted at building $b_l^{\prime}$ and sort $s_l^{\prime}$. Human
planners at a building \( b_l^{\prime} \in \mathcal{B} \) within a
cluster \( g \in \mathcal{G} \) evaluate the operational status of
their terminal and, based on their expertise and real-time changes,
decide which loads need to be shifted to different buildings or
sorts. This evaluation is guided by both load-level features, such as
the origin, planned volume, and planned arrival time, and
building-level features, including the planned work staff and planned
volume to be processed during each sort. A load is considered shifted
when its \textit{actual processing building} \( b_l\), or
\textit{actual processing sort} \( s_l\) differs from the planned
processing location, i.e. $(b_{l}, s_{l}) \neq (b^{\prime}_{l},
s^{\prime}_{l})$. There are two types of shifts:

\begin{itemize}
    \item \textbf{Internal shifts}: These occur when the actual processing building is the same as the planned processing building for load $l$, but the actual processing sort, denoted by $s_l$, differs from the planned processing sort, denoted by $s_l^{\prime}$ (i.e., $b_l^{\prime} = b_l$ and $s_l^{\prime} \neq s_l$). 
    \item \textbf{External shifts}: These occur when the actual processing building of load $l$, denoted by $b_l$, is different from the planned processing building, denoted by $b_l^{\prime}$ (i.e., $b_l^{\prime} \neq b_l$).
\end{itemize}

\begin{figure}[!t]
    \centering
    \includegraphics[width=0.45\textwidth]{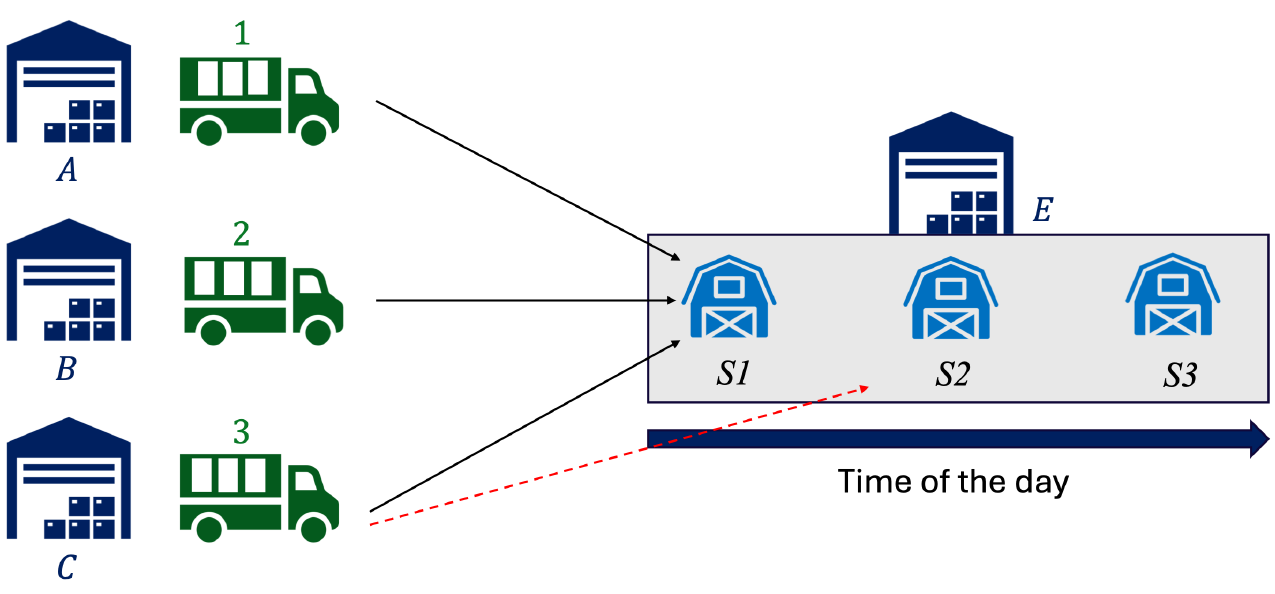}
    \caption{Illustration of an internal shift, where load $3$ was 
    planned to be processed during sort $S1$ in building $E$ but was actually processed during sort $S2$ in building $E$.}
    \label{fig:internal_shift}
\end{figure}

\begin{figure}[t!]
    \centering
    \includegraphics[width=0.45\textwidth]{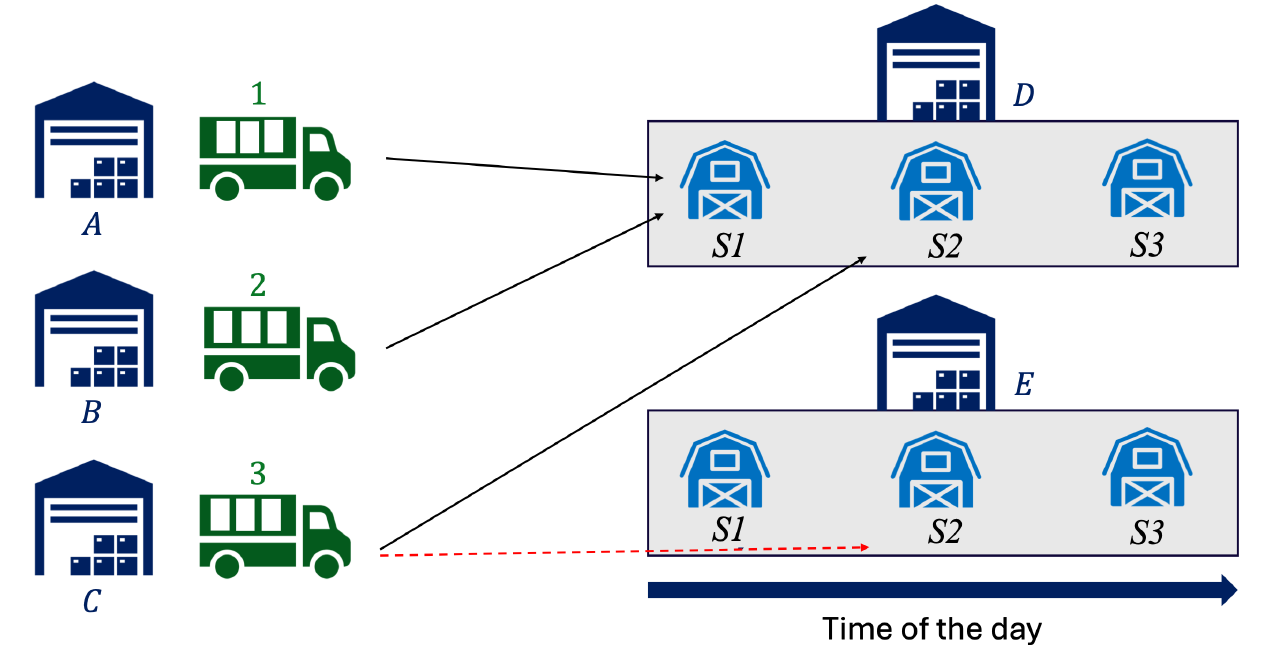}
    \caption{Illustration of an external shift, where load $3$ was planned to be processed during sort $S2$ in building $D$ but was actually processed during sort $S2$ in building $E$.}
    \label{fig:external_shift}
\end{figure}

Figure \ref{fig:internal_shift} presents an internal shift, where load
$3$, with planned processing building $b'_3 = E$ and planned
processing sort $s'_3 = D$, is processed in the planned building
(i.e., $b'_3 = b_3 = E$), but at a different sort $s_3 =
T$. Similarly, Figure \ref{fig:external_shift} depicts an an external
shift for load $3$, where the load is redirected from the planned
processing building $b'_3 = E$ to $b_3=F$. Note that the actual
processing sort for external shifts may be the same or differ from the
planned sort.

Given a set of loads $\mathcal{L}$ and their associated feature vectors $\mathbf{X}$, which includes both load-level and building-level features for the planned processing building, the objective is to provide a data-driven decision-support tool to predict the processing building and sort for each planned load (i.e., $b_l$ and $s_l$ for $l \in \mathcal{L}$, with predictions denoted as $\hat{y}_{b, w} $ for building and, $\hat{y}_{s, w}$ and $\hat{y}_{s, d}$ for sort. Here, the subscript $w$ refers to the predictions made one week prior to the day of operations while the subscript $d$ corresponds to the predictions made on the day of operations). This paper develops a confidence-aware prediction method to reduce the cognitive burden of planners and operators in tactical and operational settings.

\section{Data Overview} \label{Data Overview}

This section reviews the data available for decision support.

\begin{table*}[!t]
\centering
\caption{Description of the numerical features. The planned values refer to the operations expected to be performed at the planned processing building and sort \((b_l^{\prime}, s_l^{\prime})\) on the estimated arrival date of load $l$.}
\label{tab:numerical-feature-description}
\fontsize{10}{20}\selectfont 
\begin{tabular*}{\textwidth}{@{\extracolsep{\fill}}ll}
\hline
\textbf{Feature} & \textbf{Description} \\ \hline
Pln Volume & Volume planned to be processed \\ \hline
Pln PPH & Planned packages processed per hour \\ \hline
Pln payroll & Planned total staffing on payroll \\ \hline
Pln work staff & Planned number of employees working \\ \hline
Pln runtime & Planned total operation run time \\ \hline
Pln process rate & Planned number of packages scanned per hour \\ \hline
Pln FPH & Planned number of unloaded and sorted packages \\ \hline
Pln unload span & Planned time unload will operate \\ \hline
Load volume & Volume of the considered load \\ \hline
Load creation date & Creation date of the load in its origin location \\ \hline
Est Arr date & Estimated Arrival date of the load \\ \hline
Est Arr Time & Estimated Arrival time of the load (used for predictions in operational settings) \\ \hline
\end{tabular*}%
\end{table*}

\begin{table*}[!tb]
\centering
\caption{Description of the categorical features}
\label{tab:categorical-features}
\fontsize{10}{20}\selectfont 
\begin{tabular*}{\textwidth}{@{\extracolsep{\fill}}llll}
\hline
\textbf{Feature}                 & \textbf{Description}  & \textbf{Unique Values} \\ \hline
Pln dest cluster & Planned destination cluster of the load & 2  \\ \hline
Pln dest building $(b_l^{\prime})$   & Planned processing building of the load & 6                      \\ \hline
Pln dest sort $(s_l^{\prime})$ & Planned processing sort of the load & 3                      \\ \hline
Org building  & Origin building of the load & 329                    \\ \hline
Org sort  & Origin sort of the load & 8                      \\ \hline
\end{tabular*}
\end{table*}

\subsection{Data Description}

This research utilized historical data from the industrial
partner. The dataset consists of 250,000 planned loads, with origins
and destinations spanning two clusters across two states and three
distinct sorts (i.e., $\mid \mathcal{G} \mid = 2$, $\mid \mathcal{B}
\mid = 6$, and $\mid \mathcal{S} \mid = 3$), collected between
September 2022 and January 2024.

The dataset comprises two levels of information essential for accurate
prediction: 1) load-level data and 2) building-level data. Load-level
data contains detailed information for each planned load, including
origin, planned destination, planned volume, and estimated arrival
time. Each load is associated with a planned processing building and
sort, as well as the actual building and sort where it was
processed. The building-level data is associated with the planned
processing building on the estimated arrival date. It includes
building-level features such as the number of packages scheduled to be
processed at the location, along with the planned workforce.

The dataset features are categorized into temporal, numerical, and
categorical types. After their encoding (described in Section
\ref{Preprocessing}), temporal features are treated as numerical and
are referred to as such hereafter. The temporal and numerical features
are described together in Table
\ref{tab:numerical-feature-description}, and the categorical features
are detailed in Table \ref{tab:categorical-features}. The estimated
arrival time of the loads at the processing facility is a crucial
feature, which is used exclusively for predictions on the day of
operation (i.e., stage 2).

The dataset is first sorted based on estimated arrival times due to
its temporal nature. It is then divided into training, validation,
calibration, and testing sets (see Section \ref{Training} for further
details). The training set is used to tune the model, while the
validation set helps identify the optimal hyperparameters. The
calibration set is specifically reserved for uncertainty
quantification through conformal prediction. The test set is used for
evaluation of the model performance. The specifics of the models and
conformal prediction are detailed in Section \ref{sec: Method}
followed by the experimental details in Section \ref{Experimental
  Details}.

\begin{figure*}[!t]
    \centering
    \begin{subfigure}[b]{0.43\textwidth}
        \centering
        \includegraphics[width=\textwidth]{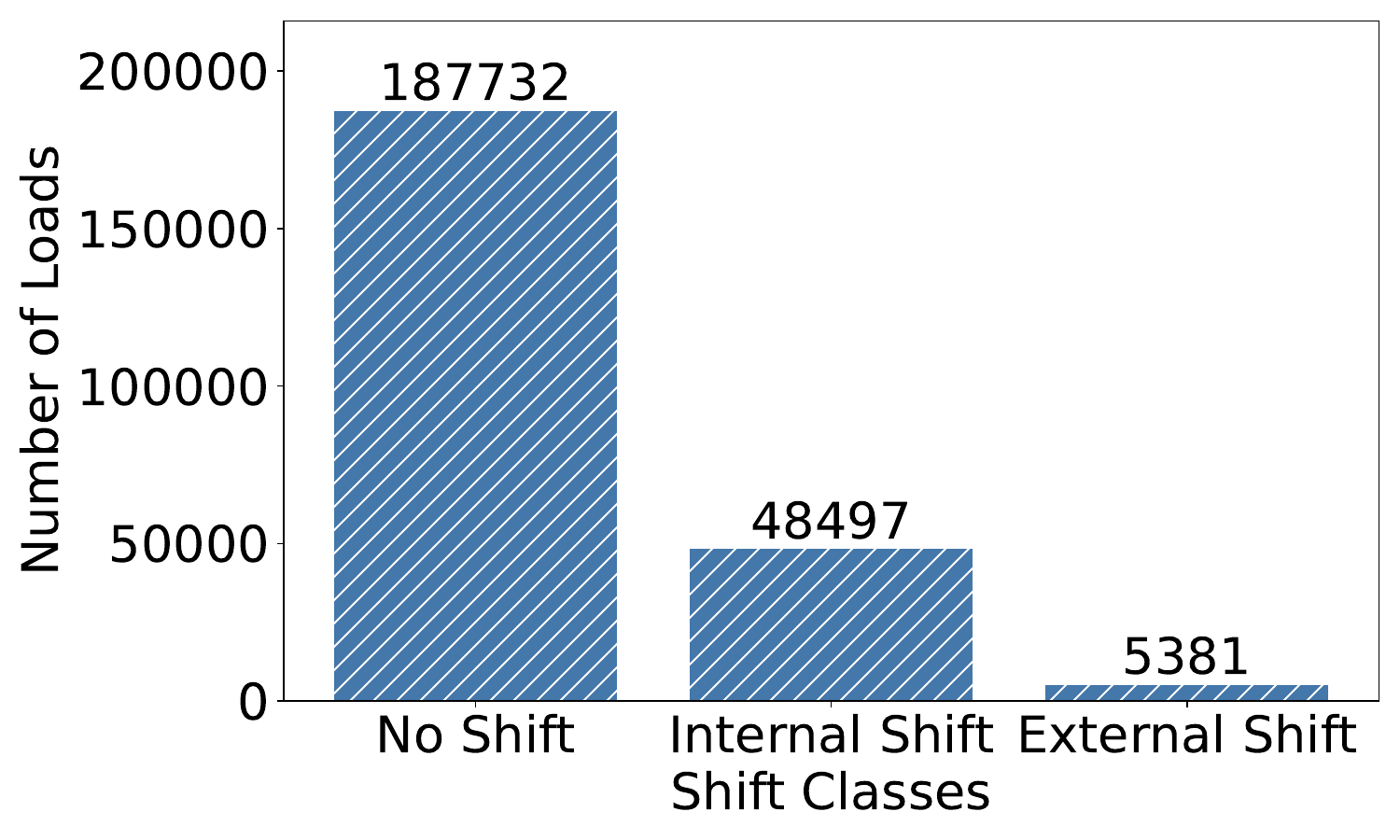}
        \caption{\centering Number of loads per shift class}
        \label{fig:analysis_shift_class}
    \end{subfigure}
    \hspace{0.05\textwidth}
    \begin{subfigure}[b]{0.43\textwidth}
        \centering
        \includegraphics[width=\textwidth]{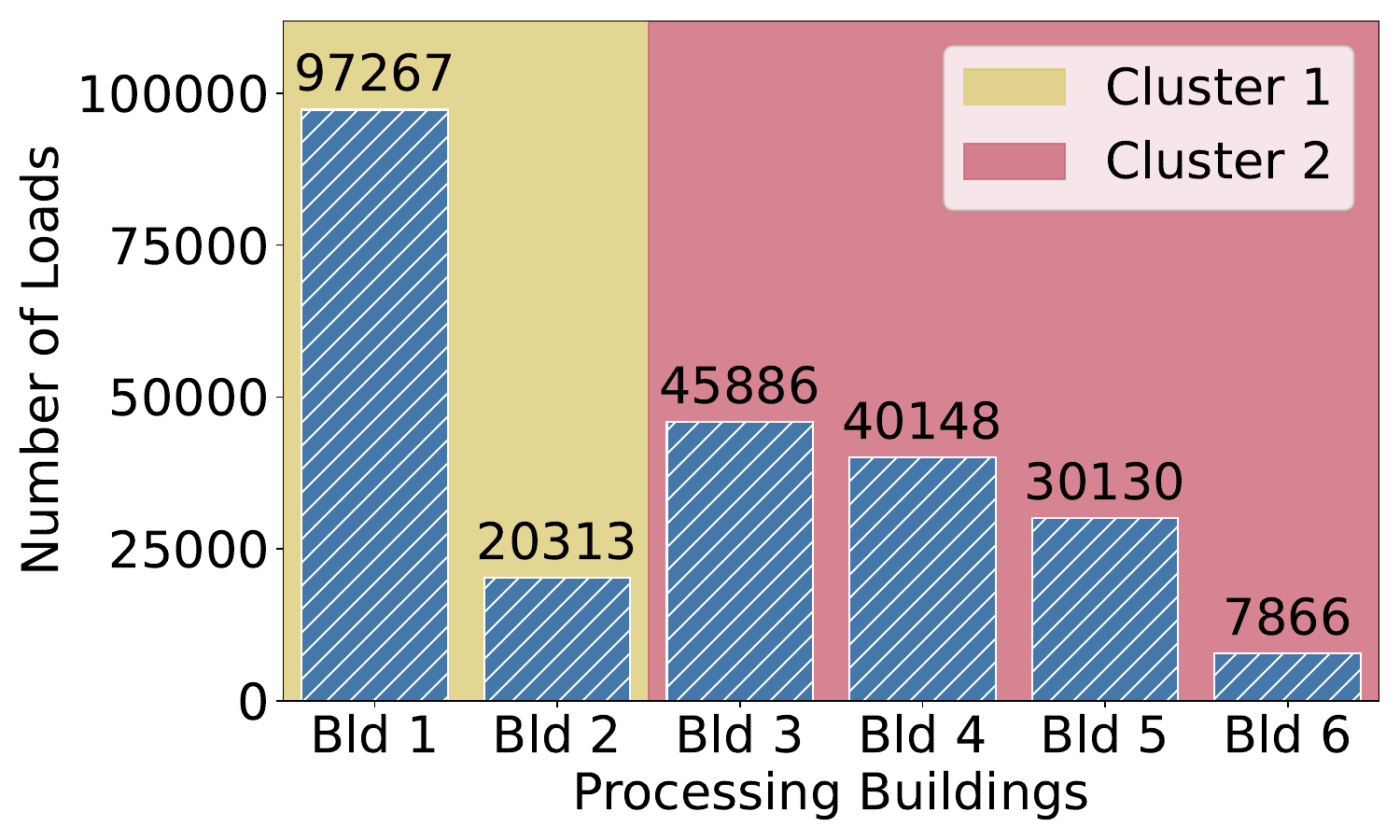}
        \caption{\centering Number of loads per building}
        \label{fig:analysis_process_bld}
    \end{subfigure}
    
    \vspace{0.05\textwidth}
    
    \begin{subfigure}[b]{0.43\textwidth}
        \centering
        \includegraphics[trim=0 10 0 3, clip, width=\textwidth]{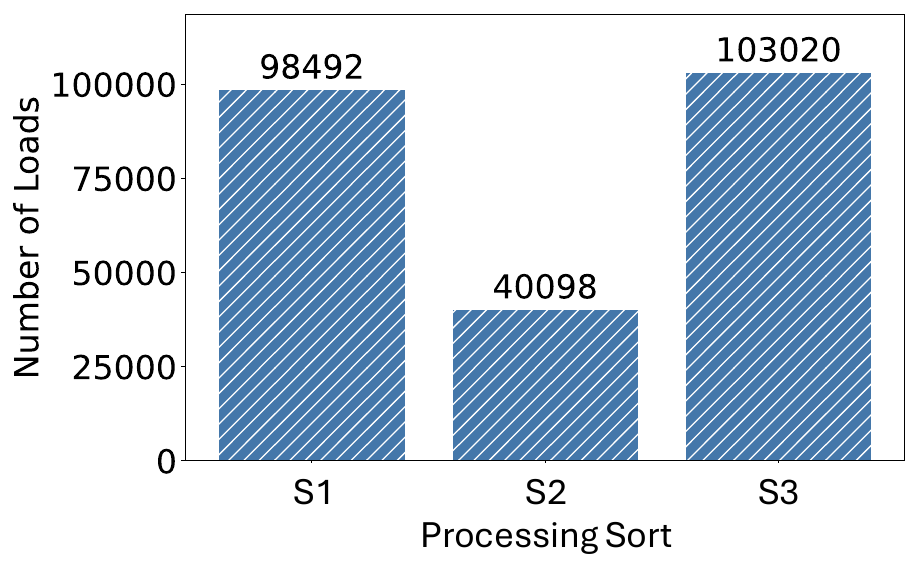}
        \caption{\centering Number of loads per sort}
        \label{fig:analysis_process_srt}
    \end{subfigure}
    \hspace{0.05\textwidth}
    \begin{subfigure}[b]{0.43\textwidth}
        \centering
        \includegraphics[width=\textwidth]{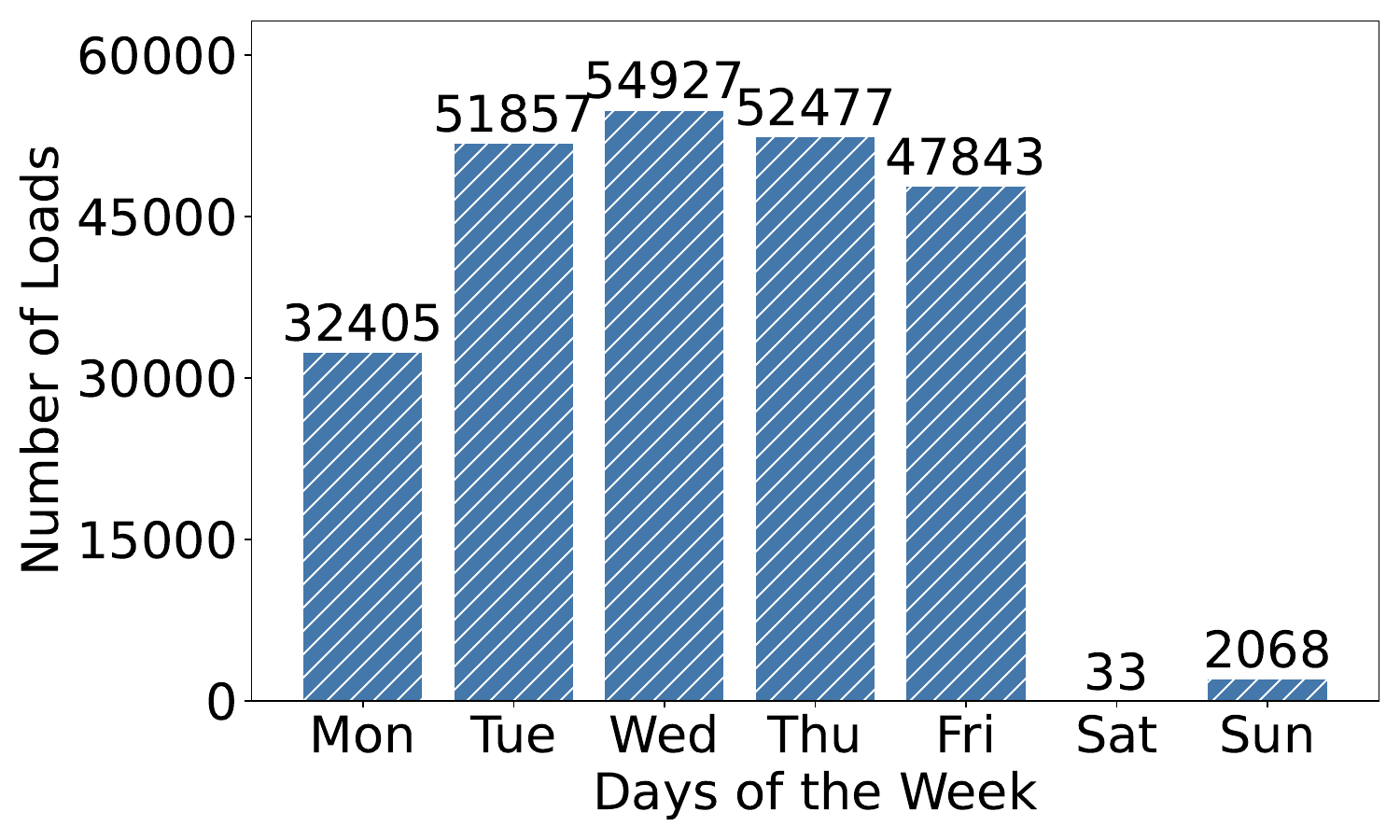}
        \caption{\centering Number of loads per day of the week}
        \label{fig:analysis_day_of_week}
    \end{subfigure}
    
    \caption{Number of loads through different dimensions: (a) The number of loads per shift class. (b) The number of loads processed in different buildings. (c) The number of loads processed in each sort. (d) The number of loads processed during each day of the week.}
    \label{fig:data_analysis}
\end{figure*}

\subsection{Data Analysis}

Figure \ref{fig:data_analysis} presents the distribution of loads
across several dimensions. Figure \ref{fig:data_analysis}(a) shows
that the data contains significant class imbalance among different
shift classes with majority of the loads coming from the "No Shift"
class, meaning that the planned processing location is same as the
processing location. {\em A crucial challenge addressed by the
  proposed framework is to identify shifted loads without generating
  excessive false alarms, i.e., predicting shifts for loads that
  should not be shifted.} Furthermore, Figures \ref{fig:data_analysis}(b) and \ref{fig:data_analysis}(c) demonstrate that the
distribution of loads across different buildings and sorts is not
uniform. Specifically, Building 1 from Cluster 1 processes around 41\%
of the loads, while the second most active building processes
approximately half as many. Regarding sorts, S2 processes only
around 17\% of the loads, while the rest of the loads is uniformly
processed by S1 and S3 sorts. Additionally, Figure \ref{fig:data_analysis}(d) reveals that the network is almost inactive during
weekends. {\em The imbalance and complexity of the data underscore the need
for a unified decision-making tool that leverages historical data from
the whole network to make informed decisions.}

\section{The Machine Learning Framework} \label{sec: Method}

Consider a set of loads $\mathcal{L}$ and their corresponding input features $\mathbf{X}$, which consists of a set of numerical and categorical features, i.e., $\mathbf{X}=(\mathbf{X}^{cont}, \mathbf{X}^{cat})$, as outlined in Tables \ref{tab:numerical-feature-description} and \ref{tab:categorical-features}. The goal is to make predictions for each load $l \in \mathcal{L}$ at two time points: 1) the building and sort one week before the day of the operation and 2) predicting the sort on the day of operation, given additional features that became available. Therefore, the three target prediction tasks can be formalized as follows:\\
\textbf{Stage 1}: Week ahead prediction:
\begin{itemize}
    \item $\mathbf{f}_{b, w}: \mathbf{X}_{b, w} \rightarrow y_b$, where $\mathbf{X}_{b, w} \subseteq \mathbf{X}$ is the set of features available one week before the day of operations for the building prediction.
    
    \item$\mathbf{f}_{s, w}: \mathbf{X}_{s, w} \rightarrow y_s$, where $\mathbf{X}_{s, w} \subseteq \mathbf{X}$ corresponds to $\mathbf{X}_{b, w}$ enriched with $y_b$ during training and $\hat{y}_b$ during inference.
\end{itemize}
\textbf{Stage 2}: Day of operation prediction:

\begin{itemize}
    \item $\mathbf{f}_{s, d}: \mathbf{X}_{s, d} \rightarrow y_s$, where $\mathbf{X}_{s, d} \subseteq \mathbf{X}$ corresponds to $\mathbf{X}_{s, w}$ enriched with the estimated arrival times of the loads.
\end{itemize}

\begin{figure*}[!t]
    \centering
    \includegraphics[width=1\textwidth]{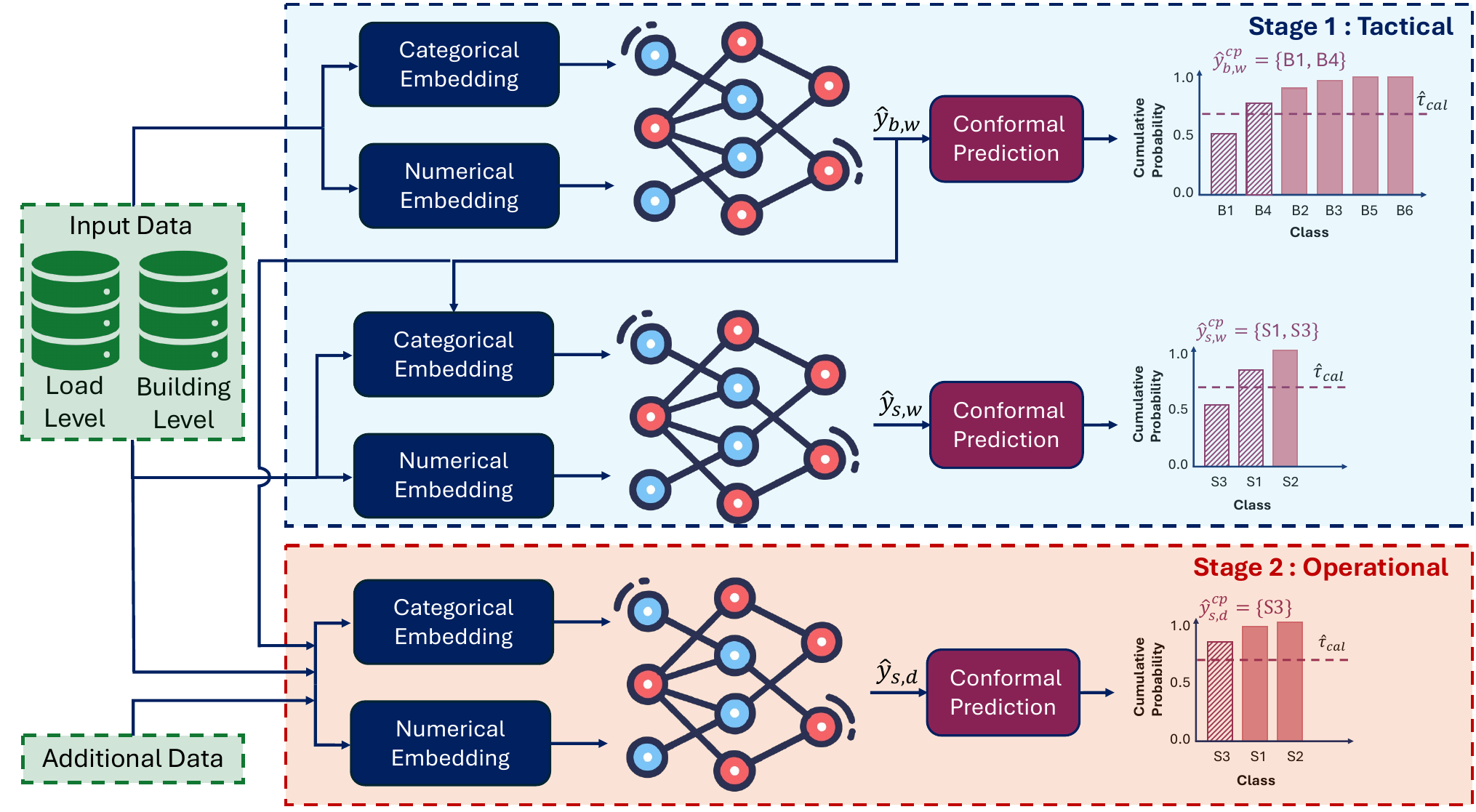}
    \caption{The Framework description. First, the data is embedded through dedicated layers for categorical and numerical features respectively. Then, a neural network predicts the building of the load. This prediction is used as an input feature for the sort prediction, which has a similar architecture. The sort prediction is refined with additional information on the day of operations (i.e., Stage 2). Conformal prediction is performed for both building and sort predictions to provide prediction sets with a pre-specified confidence level. The figure illustrates the cumulative probabilities with classes ordered in decreasing order of estimated probabilities.}
    \label{fig:architecture}
\end{figure*}

As discussed in Section \ref{pbm_statement}, the motivation behind the
two-step decision-making process is to predict the buildings and sorts
for the tactical planing, and use the most recent status of the loads
to predict the sorts on the day of operation and adjust to the
uncertainties in the network for operational planing. Each one of
these mappings is learned through supervised empirical risk
minimization. By omitting the subscripts (for simplicity and without
loss of generality), the learning problems can be formalized as:
\begin{equation} \label{eq: risk minimization}
    \min_\mathbf{\theta} \, \mathbb{E}_{(\mathbf{X}, y) \sim P_{\mathcal{D}}} [L(\mathbf{f}(\mathbf{X}, \mathbf{\theta}), y)],
\end{equation}
where $\mathbf{\theta}$ is the vector of model parameters,
$P_{\mathcal{D}}$ is the data distribution, and $L$ is the loss
function (e.g., categorical cross entropy loss). Since the true data
distribution $P_\mathcal{D}$ is usually unknown, empirical risk
minimization (ERM) is performed to find the model parameters, using
$N$ pairs of training data $\mathcal{D}_\text{train}$, where $N =
|\mathcal{D}_\text{train}|$.

Figure \ref{fig:architecture} shows the schematic of the proposed DL
architecture. During the training process, the building prediction
model $\mathbf{f}_{b, w}$ takes as input the feature vector
$\mathbf{X}_{b, w}$, which corresponds to the features available a
week before the day of operation. Next, the numerical and categorical
features are mapped into a new vector space through dedicated
embedding layers to capture the complex and nonlinear relations in the
structured data. The resulting data is then passed into the Neural
Network (NN) (e.g., multi-layer perceptron (MLP) or Residual Network
(ResNet)) and the model parameters are optimized by minimizing the
categorical cross entropy loss. The tactical-level sort prediction
model $\mathbf{f}_{s, w}$ uses input features similar to the building
prediction model but also includes the actual processing buildings as
an additional feature (i.e., $\mathbf{X}_{s, w}$), which is essential
for accurate sort predictions. Finally, the sort prediction model for
the day of operation, i.e., $\mathbf{f}_{s, d}$, is trained using the
same input features as model $\mathbf{f}_{s, w}$ plus the estimated
arrival time of the loads as an additional feature (i.e.,
$\mathbf{X}_{s, d}$). Each of these models is trained using the same
training dataset $\mathcal{D}_\text{train}$ and the hyperparameters
are tuned using a validation dataset
$\mathcal{D}_\text{validation}$. A calibration dataset
$\mathcal{D}_\text{calibration}$ is used for conformal prediction, to
provide a prediction set that guarantees to include the correct label
with a pre-defined confidence level.

At inference time, the processing building is first predicted as
$\hat{\mathbf{f}}_{b, w}(\mathbf{X}_{b, w}) = \hat{y}_{b, w}$ along
with the conformalized prediction set $\hat{y}^{cp}_{b, w}$. Next, the
predicted building $\hat{y}_{b, w}$ (i.e., output class of the softmax
with the highest probability) is concatenated with the feature vectors
and is used as an input to the week-ahead sort prediction model to
predict the processing sort as $\hat{\mathbf{f}}_{s, w}(\mathbf{X}_{s,
  w}) = \hat{y}_{s, w}$ and the prediction set $\hat{y}^{cp}_{s,
  w}$. Finally, on the day of operation, the trained model
$\hat{\mathbf{f}}_{s, d}$ takes as input the feature vectors
$\mathbf{X}_{s, d}$ concatenated with the building prediction
$\hat{y}_{b, w}$ and the estimated arrival time of the load as
additional features to predict the final processing sort as
$\hat{y}_{s, d}$ and the prediction set $\hat{y}^{cp}_{s,
  d}$. Sections \ref{method - feature emb} and \ref{Method - Conformal
  Prediction} provide further details on the embedding layers and the
conformal prediction model.

\subsection{Feature Embedding} \label{method - feature emb}

Embeddings play a crucial role in enhancing the performance of DL
models over tabular and unstructured data as they allow models to more
effectively capture the underlying relationships within the data
\citep{gorishniy2023embeddingsnumericalfeaturestabular}. The proposed
framework employs dedicated embeddings to handle categorical and
numerical features separately.

\subsubsection{Categorical Embeddings\\} \label{sec: categorical embed}
\begin{figure}[tb]
    \centering
    \includegraphics[width=0.4\textwidth]{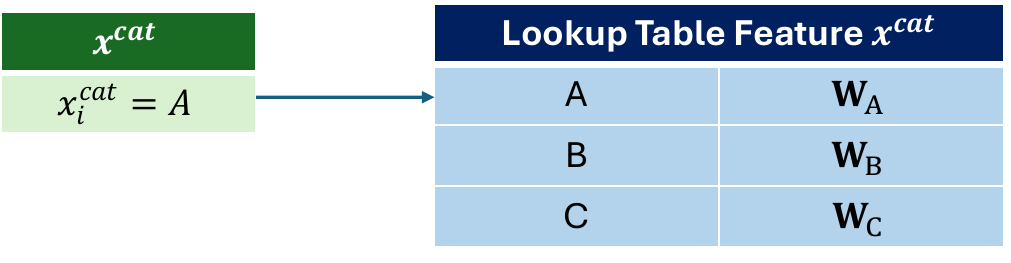}
    \caption{Categorical Embeddings: Value $x^{cat}_i = A$ of feature $x^{cat}$ for sample $i$ is embedded as vector $W_A$ using the Look-up table of the corresponding feature.}
    \label{fig:categorical_embedd}
\end{figure}

A powerful technique for embedding categorical features, which is
mainly popular within natural language processing domain for
generation of word embeddings, amounts to creating a trainable look-up
table that maps each unique value of a categorical feature to an
$n$-dimensional vector
\citep{mikolov2013efficientestimationwordrepresentations}. The process
begins by initializing the look-up table randomly. The entries are
then adjusted iteratively through backpropagation to optimize the
embeddings. Figure \ref{fig:categorical_embedd} shows a simple example
of a look-up table for a categorical feature.

Consider a categorical feature $x^{cat} \in \mathbf{X}^{cat}$ with $C$
unique values. Each unique value $u$ of $x^{cat}$ is mapped to a
$n$-dimensional vector $\mathbf{W}_u$, forming a $(C \times n)$
look-up table, where $C$ represents the cardinality of the
feature. In this paper, the embedding vector dimension $n$ can be specified as:
\begin{equation}
    n = \min(n_{max}, \frac{C+1}{2}) \label{size of cat embeddings},
\end{equation}
where $n_{max}=50$ is the maximum size of the embedding for the
categorical features. This embedding approach allows the model to
learn dense vector representations of categorical variables, capturing
complex relationships that enhance performance in downstream
tasks. The embeddings are trained separately for building and sort
prediction, as shown in Figure \ref{fig:architecture}.

\subsubsection{Numerical Embeddings\\} \label{numerical Embeddings}

Embedding numerical features can also enhance the performance and
robustness of DL models significantly on tabular datasets
\citep{gorishniy2023embeddingsnumericalfeaturestabular}. Specifically,
Gorishniy et al. introduced two embedding techniques: Quantile-based
Piecewise-Linear Embeddings (QL) and Periodic Linear Embeddings
(PLR). Both methods have demonstrated strong performance, as reported
by \cite{gorishniy2023embeddingsnumericalfeaturestabular}, and
extensive experiments have been performed to evaluate their
performance in this paper.

\begin{itemize}
    \item \textbf{QL Embedding}
\end{itemize}

\begin{figure}[tb]
    \centering
    \includegraphics[width=0.47\textwidth]{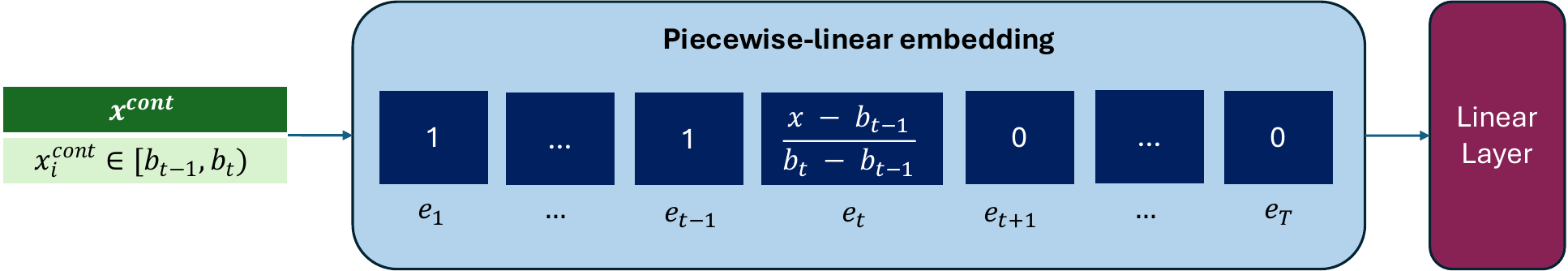}
    \caption{QL Embeddings : Quantile-based bins are created based on the distribution of the values of the numerical features. Then, values are encoded individually based on the the bin they are in. A linear layer is added after the encoding step.}
    \label{fig:QL}
\end{figure}

Given a numerical feature $x^{cont} \in \mathbf{X}^{cont}$, this feature is mapped into a $T$-dimensional vector as illustrated in Figure \ref{fig:QL}. Formally, for a sample data point $i$, the embeddings are defined as follows : 

\begin{equation}
    \mathbf{z}_i = f(x^{cont}_i) = \text{Linear}(\text{PLE}(x^{cont}_i)),
\end{equation}

where : 

\begin{equation}
    \begin{aligned}
        PLE(x) &= [e_1,\ \dots,\ e_T] \in \mathbb{R}^T, \\
        e_t &=
        \begin{cases}
            0 & \text{if } x < b_{t - 1}\ \text{and}\ t > 1,\\
            1 & \text{if } x \ge b_t\ \text{and}\ t < T,\\
            \frac{x - b_{t - 1}}{b_t - b_{t - 1}} & \text{otherwise},
        \end{cases}
    \end{aligned}
\end{equation}
and $[b_{t-1}; b_t)$ are bins obtained from empirical quantiles for
  each individual feature. For feature $x^{cont} \in
  \mathbf{X}^{cont}$, $b_t = Q_{\frac{t}{T}}(\{x^{cont}_{i}\}_{i \in
    \mathcal{D}_{train}})$ where $Q_{\frac{t}{T}}$ is the empirical
  quantile function with respect to the ${\frac{t}{T}}$
  percentile. The number of bins $T$ is fixed for the different
  numerical features. Additionally, the linear layer applies an affine
  transformation to the input. Formally, for an input vector
  $\mathbf{x}$, the output of the Linear layer is computed as
  $\mathbf{y} = W\mathbf{x} + \mathbf{b}$, where $W$ is a weight
  matrix, and $\mathbf{b}$ is a bias vector. Intuitively, these
  embeddings can be seen as an adaptation of One-Hot encoding for
  numerical features, further enhanced by a linear layer.

\begin{itemize}
    \item \textbf{PLR embeddings}
\end{itemize}

\begin{figure}[tb]
    \centering
    \includegraphics[width=0.49\textwidth]{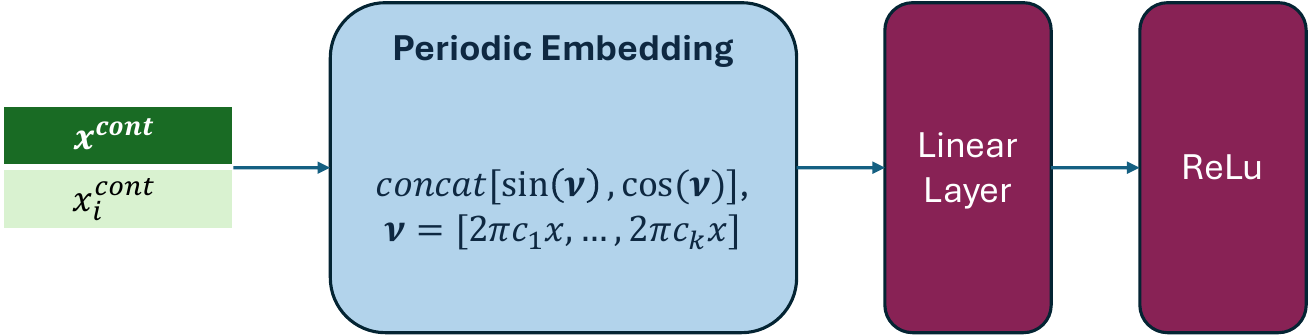}
    \caption{PLR Embeddings : Numerical features are first embedded using periodic embeddings. Then, a linear layer and ReLu activation functions are applied to the output.}
    \label{fig:PLR}
\end{figure}

These embeddings, inspired by Fourier transformations, were first
introduced by \cite{tancik2020fourierfeaturesletnetworks} and further
enhanced by \cite{gorishniy2023embeddingsnumericalfeaturestabular} for
applications on tabular data.  A PLR embedding first maps a numerical
feature $x^{cont} \in \mathbf{X}^{cont}$ into a $2k$-dimensional
vector using a periodic function, where $k$ is a tunable
hyperparameter. Then, a linear layer maps the obtained vector to a
$d$-dimensional vector, where $d$ denotes the dimension of the
embedding. Finally, the resulting vector is passed to a ReLU
activation function. This process is illustrated in \ref{fig:PLR} and
detailed below:

\begin{equation} 
\begin{split}
    \mathbf{z}_i &= f(x^{cont}_i) \\
                 &= \text{ReLU}(\text{Linear}(\text{Periodic}(x^{cont}_i)))
\end{split}
\label{PLR}
\end{equation}
where
\begin{equation}
    \begin{aligned}
        \text{Periodic}(x) &= \text{concat}[\sin(\mathbf{v}), \cos(\mathbf{v})], \\
        \mathbf{v} &= [2\pi c_1 x, \ldots, 2\pi c_k x],
    \end{aligned}
\end{equation}
and $c_i$ (with $i \in \{1, \ldots, k\}$) are trainable parameters.
Both embedding dimensions, $d$ and $k$, are tunable
hyperparameters.

The QL embeddings could be viewed as feature encoding since the bins
are determined by the distribution of the features in the training
set. In contrast, PLR embeddings enable end-to-end training of the
$c_i$ parameters through back-propagation. Following the numerical and
categorical embeddings, a NN model is used to predict the process
building and sorts as presented in Figure \ref{fig:architecture}. This
paper considers MLP and ResNet with and without the embedding layers
in the experiments. Additionally, traditional ML methods were used as
baselines for comparison with the proposed DL framework. Further
details are provided in Section \ref{Experimental Details}.

\subsection{Conformal Prediction} \label{Method - Conformal Prediction}

As discussed previously, conformal prediction is employed as a
post-processing step after training the models for uncertainty
quantification of the prediction. Conformal prediction is a
model-agnostic and distribution-free method that provides theoretical
guarantees on the coverage of the prediction set \citep{vovk}. Instead
of making single-point prediction, this approach utilizes a dataset
$\mathcal{D}_{calibration}$ (not seen during the training) to
calibrate the model output and generate a set of possible outcomes
that meet a user-specified target coverage rate. This confidence-aware
method is crucial for decision making in load planning, as it assists
planners by offering a set of potential processing buildings and
sorts, thereby reflecting the inherent prediction uncertainty.

\textit{Coverage} and \textit{efficiency} are the two key metrics for
evaluating uncertainty quantification in the conformal prediction
setting. Coverage represents the probability that the true label is
included in the prediction set, while efficiency refers to the average
size of the prediction set. A model is considered more reliable when
it achieves the target coverage with minimal efficiency (i.e., smaller
prediction set sizes). To balance these two metrics, this paper
employs the Regularized Adaptive Prediction Sets (RAPS) algorithm,
which includes a regularization term to penalize large prediction sets
\citep{angelopoulos2022uncertaintysetsimageclassifiers}. Given a
trained classification model $\mathbf{\hat{f}}$ and a calibration set
$\mathcal{D}_{calibration}$, the goal is to learn a function
$\mathcal{C}$ that forms prediction sets containing the true label in
the test set. More precisely, the goal is to find a function
$\mathcal{C}(.)$ that, given an input $X_{test}$, computes a
prediction set $\mathcal{C}(X_{test})$ that contains the ground truth
$y$ with a probability satisfying
\begin{equation} \label{eq: cp_ineq}
    1- \alpha \leq \mathbb{P}(y \in \mathcal{C}(X_{test}))
\end{equation}
where $1 - \alpha$ is the desired target coverage with $\alpha \in [0,
  1]$.

It is important to note that one can satisfy Equation \eqref{eq: cp_ineq} and reach the target coverage by simply considering a large
prediction set size, which is not desirable in practice. Hence, the design of the function $\mathcal{C}$ is crucial for providing high-quality solutions that achieve the target coverage with optimal
efficiency (i.e., relatively small average prediction set size). In
general, conformal prediction consists of two different steps: 1) the
calibration step; and 2) the generation of the prediction
sets. \cite{angelopoulos2022uncertaintysetsimageclassifiers} proposed
the following procedure for these steps, with the calibration step
computing an upper probability threshold $\hat{\tau}_{cal}$ and the generation
step uses $\hat{\tau}_{cal}$ to compute the prediction set.

During the calibration phase, for each sample $i \in
\mathcal{D}_{calibration}$ associated with a feature vector
$\mathbf{X}$, the class labels are sorted from the most probable one
to the least probable one according to the output probabilities of the
pre-trained model $\mathbf{\hat{f}}$. Then, given that $y \in
\mathcal{Y}$ (where $\mathcal{Y} = [1;K]$ and $K$ is the number of
classes) is the true label, the following score is calculated as:

\begin{equation}
\begin{split}
    E_i &= \sum_{y^{\prime}=1}^{K} \hat{\pi}_{x}(y^{\prime}) 
    \mathbb{I}_{\{ \mathbf{\hat{\pi}_x}(y') \geq \mathbf{\hat{\pi}_x}(y) \}} \\
    &\quad + \lambda (o_x(y) - k_{reg})^{+}
\end{split}
\end{equation}

where $\hat{\pi}_{x}(y^{\prime})$ is the output probability given by
$\mathbf{\hat{f}}$ for sample $i$ and class $y^{\prime}$, and
$o_\mathbf{x}(y) = \mid \{ y' \in \mathcal{Y} :
\mathbf{\hat{\pi}_x}(y') \geq \mathbf{\hat{\pi}_x}(y)\}\mid$
represents the rank of label $y$ among all possible labels, as
determined by the probability estimate $\mathbf{\hat{\pi}_x}$ (e.g.,
if y is the third most likely label, $o_\mathbf{x}(y) = 3$). The
$(.)^+$ operator denotes the positive part. $\lambda$ and $k_{reg}$
are regularization hyperparameters : $k_{reg}$ is the size of the
prediction set from which the algorithm starts penalizing additional
labels and $\lambda$ is the associated penalty factor. More
intuitively, $E_i$ corresponds to the cumulative estimated
probabilities of the labels until the true label plus a regularization
term that penalizes large prediction sets. Based on these
calculations, the output $\hat{\tau}_{cal}$ is calculated as the
$\lceil (1-\alpha)(1+n) \rceil$ largest value in $\{E_i\}_{i=1}^{n}$
with $n = \mid \mathcal{D}_{calibration}\mid$; it can be thought as
the score that acheives the target coverage and not more.

At inference time, using $\hat{\tau}_{cal}$ calculated during the
calibration step, the goal is to create the prediction set for a
sample $j \in \mathcal{D}_{test}$. Without loss of generality, it is
assumed that the output labels $\mathcal{Y} = [1;K]$ are sorted from
the most to the least probable in terms of estimated probabilities for
sample $j$. Then, the size of the prediction set for sample $j$ is defined as:

\begin{equation}
\begin{split}
    M &= \mid \{ p \in \mathcal{Y} : \sum_{y^{\prime}=1}^{p}\hat{\pi}_{x}(y^{\prime}) \\
    &\quad\quad\quad\quad\quad + \lambda(j-k_{reg})^{+} \leq \hat{\tau}_{cal} \} \mid + 1
\end{split}
\end{equation}

It represents the number of labels to be included in the prediction
set, while not exceeded the threshold constraint given by $\tau_{cal}$
(the term $+1$ is added to disallow empty prediction sets). The $M$
most probable classes form the $1-\alpha$ confidence set.

This study uses RAPS as a post-processing step for building and sort
predictions
\citep{angelopoulos2022uncertaintysetsimageclassifiers}. The
parameters $\lambda$ and $k_{reg}$ are optimized using an ad-hoc
strategy. Specifically, $\lambda$ is set to 0.001 and $k_{reg}$ to 2, as
these values provided a good trade-off between achieving high coverage
and maintaining a small average prediction set size.

\section{Experimental Details} \label{Experimental Details}

This section presents the experiments that use the dataset described
in Section \ref{Data Overview}. These experiments compare traditional
ML methods with the proposed models. Specifically, the baseline models
include Decision Tree (DT), Random Forest (RF), LightGBM (LGBM),
CatBoost, and XGBoost. For the DL-based models, both MLP and ResNet
architectures are considered. Categorical embeddings are applied in
all DL-based experiments, while the performance is evaluated with and
without the inclusion of numerical embedding layers.

\subsection{Preprocessing} \label{Preprocessing}

As detailed in Section \ref{Data Overview}, the dataset consists of
numerical, categorical, and temporal features. A preprocessing step is
performed before feeding the data into the models as outlined below:

\subsubsection{Numerical Features} For the traditional ML models, the
numerical features are used without any modification. However, for the
DL-based models, a Quantile Transformer is used following the addition
of noise (distributed as $\mathcal{N}(0, 10^{-5})$) to normalize the
data and facilitate the training process.

\subsubsection{Categorical Features} For traditional ML models (excluding
CatBoost), label encoding is applied to features with low cardinality
(less than 20 categories). Label encoding assigns a unique integer to
each category. Target encoding, which replaces categories with the
mean of the target variable, is used for high-cardinality
features. CatBoost is trained with its internal encoder for
categorical features. For the DL-based models, all categorical
features are initially label-encoded and then passed though the
categorical embedding layer, as detailed in Section \ref{sec:
  categorical embed}.

\subsubsection{Temporal Features} These features, corresponding to the
event dates (e.g., scheduled processing date of the load), are first
decomposed into day, week, and month components and are
label-encoded. Next, given their periodic nature, cyclical encoding is
applied to capture continuity (e.g., January is adjacent to December,
and Monday is close to Sunday). Each cyclical feature \( g \) is
represented by two encoded features:
\begin{equation}
    g^{sin} = \sin\left(\frac{2\pi g}{\max(g)}\right)
\end{equation}
\begin{equation}
    g^{cos} = \cos\left(\frac{2\pi g}{\max(g)}\right)
\end{equation}
Following the preprocessing and encoding steps described above, the
data is then used as inputs to the models for training.

\subsection{Training}\label{Training}

\subsubsection{Data Split\\}
\begin{figure}[tb]
    \centering    \includegraphics[width=\linewidth]{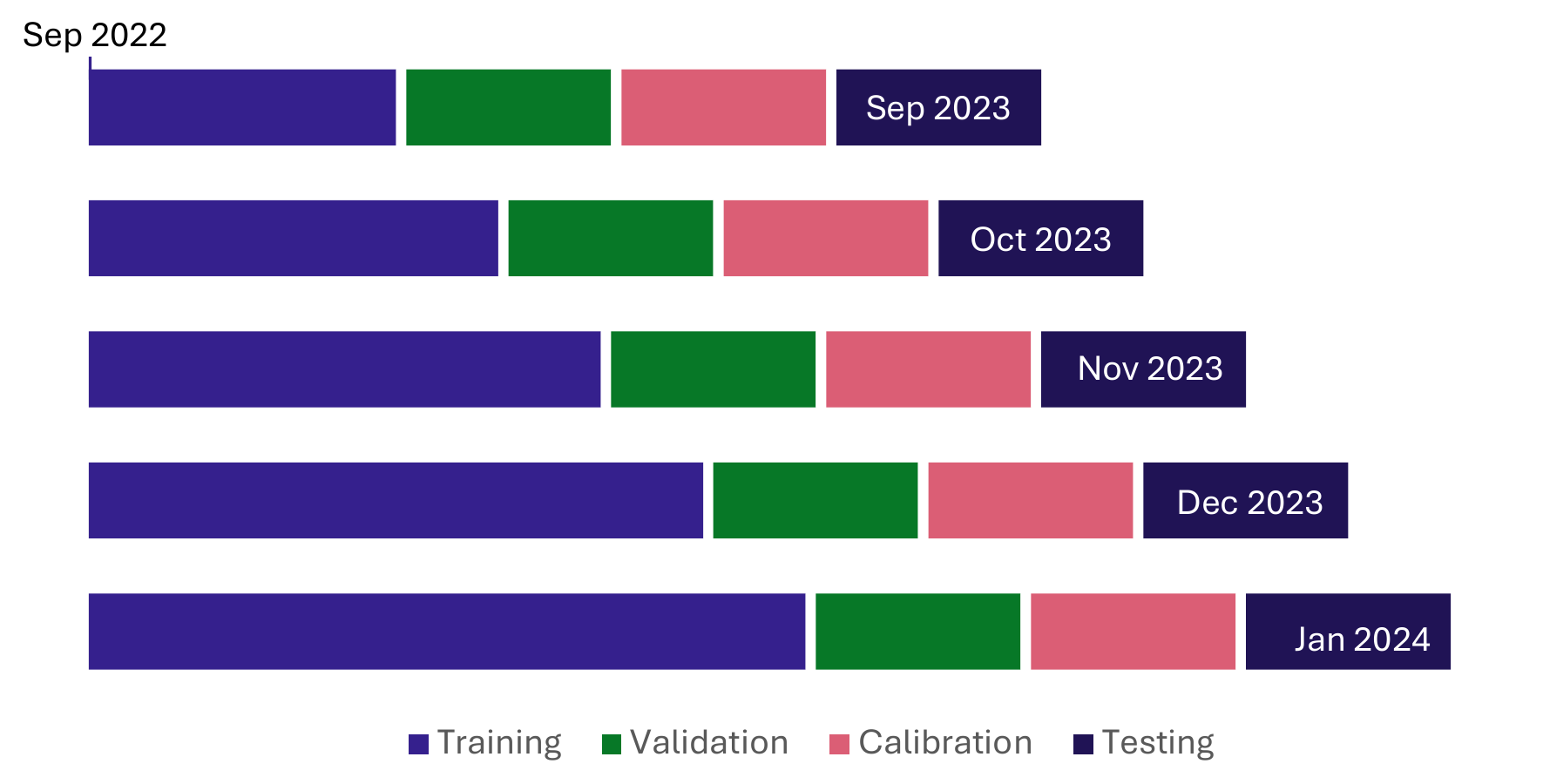}
    \caption{Data splits for experiments : The data was splitted temporaly in training, validation, calibration and testing sets. Five experiments were conducted with different testing sets.}
    \label{fig:data-splits}
\end{figure}

The experiments are conducted across five distinct time horizons, as shown in Figure \ref{fig:data-splits}, to compare models and evaluate their consistency over time. Given the temporal natural of the data, it is first ordered according to the estimated arrival date of the loads and the last month is kept for testing. From the remaining data, 10\% is allocated to the calibration set for conformal prediction , 10\% to the validation set for hyperparameter, and 80\% to the training set. All the experiments are repeated with the same strategy and the average and standard deviation of the metrics are calculated. The results are provided in Section \ref{sec : Results}.

\subsubsection{Hyperparameter tuning\\}
Optuna, an efficient library for hyperparameter tuning \citep{akiba2019optunanextgenerationhyperparameteroptimization}, is used with 100 trials for the traditional ML models. For the DL-based models, grid search is used. The range of parameters for each model are provided in Appendix \ref{appendix}.

\subsubsection{Experimental Set up\\}
The experiments are conducted using 1 NVIDIA RTX 6000 GPU and 6 CPUs per task. Each GPU is allocated 16GB of memory. Scikit-Learn \cite{pedregosa2018scikitlearnmachinelearningpython} is used for the implementaion and training of the traditional ML models. The training of the DL models is performed using the Adam optimizer and Cross Entropy loss in PyTorch \cite{paszke2019pytorchimperativestylehighperformance}. The number of epochs is set to 20 with a patience of 5 across all experiments.

\section{Results} \label{sec : Results}

This section presents the numerical results comparing different
methods as discussed in Section \ref{Experimental Details}. The
experiments are conducted across five distinct time horizons with the
reported metrics representing mean ± standard deviation. A detailed
explanation of the time horizons is provided in Section
\ref{Experimental Details}. Metrics under the 'All Data' category
reflect the accuracy across the entire testing dataset, while the 'No
Shift', 'Internal Shift', and 'External Shift' categories represent
the accuracy specific to each respective shift class.

\subsection{Building Prediction}

\begin{table*}[t!]
    \centering
    \caption{Comparison of the accuracies for the Building Prediction task: The column labeled 'All Data' represents the accuracies on the complete testing set, while the remaining columns display accuracies based on a partition of the testing set corresponding to different shift classes of the loads.}
    \renewcommand{\arraystretch}{1.3} 
    \resizebox{0.9\textwidth}{!}{
        \begin{tabular}{c c c c c c}
            \hline
            \multirow{2}{*}{\textbf{Model}} & \multirow{2}{*}{\textbf{Num Emb}} & \multicolumn{4}{c}{\textbf{Building Prediction Accuracy}} \\
            \cline{3-6}
            & & \textbf{All Data} & \textbf{No Shift} & \textbf{Internal Shift} & \textbf{External Shift} \\
            \hline
            DT & - & 0.988 ± 0.003 & 0.996 ± 0.001 & 0.994 ± 0.002 & 0.587 ± 0.031 \\
            RF & - & 0.990 ± 0.003 & 0.999 ± 0.001 & 0.998 ± 0.001 & 0.502 ± 0.039 \\
            CatBoost & - & 0.974 ± 0.035 & 0.982 ± 0.035 & 0.984 ± 0.032 & 0.529 ± 0.069 \\
            LGBM & - & 0.986 ± 0.015 & 0.995 ± 0.009 & 0.995 ± 0.008 & 0.571 ± 0.152 \\
            XGBoost & - & 0.992 ± 0.001 & 0.999 ± 0.001 & 0.998 ± 0.002 & 0.643 ± 0.030 \\
            \hline
            MLP & \color{myred}{\xmark} & 0.952 ± 0.075 & 0.960 ± 0.075 & 0.959 ± 0.075 & 0.514 ± 0.068 \\
            MLP & PLR & 0.990 ± 0.001 & 0.997 ± 0.001 & 0.997 ± 0.002 & 0.592 ± 0.058 \\
            MLP & QL & 0.990 ± 0.002 & 0.997 ± 0.001 & 0.997 ± 0.001 & 0.622 ± 0.030 \\
            ResNet & \color{myred}{\xmark} & 0.974 ± 0.030 & 0.983 ± 0.029 & 0.978 ± 0.037 & 0.533 ± 0.079 \\
            ResNet & PLR & 0.989 ± 0.002 & 0.996 ± 0.001 & 0.996 ± 0.002 & 0.590 ± 0.055 \\
            ResNet & QL & 0.990 ± 0.002 & 0.997 ± 0.001 & 0.997 ± 0.001 & 0.651 ± 0.036 \\
            \hline
        \end{tabular}
    }
    \label{tab:bldg_accuracy}
\end{table*}

Table \ref{tab:bldg_accuracy} presents the performance of different models
for building predictions under the first stage of the decision making,
which occurs one week ahead of the day of operation. Considering the
overall accuracy, XGBoost and RF predict the target processing
buildings with more than 99$\%$ accuracy. Similarly, the MLP and
ResNet models achieve similar performance with the inclusion of the
numerical embedding layers. More precisely, the presence of the
numerical embedding layers increases the overall accuracy in average
by 3.8$\%$ and 1.6$\%$ for the MLP and ResNet, respectively.

The results for the No Shift and Internal Shift classes show that the
XGBoost, RF, and DL-based models with embedding consistently achieve
over 99\% accuracy.. Similar to the overall accuracy, the effect of
the numerical embeddings is evident in enhancing the performance of
the DL-based models across these shift classes. The high accuracy in
these shift classes is essential for tactical planning, as the
majority of the data falls within the No Shift or Internal Shift
categories, making a low false alarm rate critical.

As detailed in Section \ref{Data Overview}, the dataset exhibits a
significant class imbalance, with External Shifts comprising only 2\%
of the dataset. This imbalance presents a key challenge: maximizing
overall accuracy across the entire test set while effectively
predicting External Shifts, where the processing building differs from
the intended one. This is critical for tactical load-planning where
the loads are supposed to be re-routed into another building in
advance. The results in Table \ref{tab:bldg_accuracy} under the
'External Shifts' sub-column indicate that all models achieve accuracy
between 50\% and 65\% in predicting the correct buildings within this
challenging class. Among them, ResNet-QL outperforms all other models
in this subclass, surpassing XGBoost. Additionally, the importance of
numerical embedding is evident, with accuracy improvements of 11\% and
12\% for MLP and ResNet, respectively.

Additionally, although not explicitly presented in Table
\ref{tab:bldg_accuracy}, an intriguing observation is the remarkably
low number of misclassifications between distinct clusters in the
building prediction task. Specifically, less than 0.1\% of the loads
that were processed in Cluster 1 are incorrectly predicted to be
processed in Cluster 2, or vice versa. This finding highlights the
model ability to accurately capture the interactions within clusters,
which is crucial for tactical planning, ensuring that loads are not
mistakenly shifted to buildings in a different cluster than originally
planned.

\begin{table*}[t!]
    \centering
    \caption{Comparison of the accuracies for the Sort Prediction task one week before the day of operations: The column labeled 'All Data' represents the accuracies on the complete testing set, while the remaining columns display accuracies based on a partition of the testing set corresponding to different shift classes of the loads.}
    \renewcommand{\arraystretch}{1.3} 
    \resizebox{0.9\textwidth}{!}{
        \begin{tabular}{c c c c c c}
            \hline
            \multirow{2}{*}{\textbf{Model}} & \multirow{2}{*}{\textbf{Num Emb}} & \multicolumn{4}{c}{\textbf{Sort Prediction Accuracy - Week ahead}} \\
            \cline{3-6}
             & & \textbf{All Data} & \textbf{No Shift} & \textbf{Internal Shift} & \textbf{External Shift} \\
            \hline
            DT & - & 0.801 ± 0.058 & 0.888 ± 0.078 & 0.459 ± 0.032 & 0.490 ± 0.050 \\
            RF & - & 0.824 ± 0.055 & 0.920 ± 0.080 & 0.449 ± 0.029 & 0.535 ± 0.048 \\
            CatBoost & - & 0.812 ± 0.058 & 0.903 ± 0.079 & 0.447 ± 0.043 & 0.542 ± 0.066 \\
            LGBM & - & 0.860 ± 0.016 & 0.947 ± 0.017 & 0.518 ± 0.039 & 0.587 ± 0.082 \\
            XGBoost & - & 0.862 ± 0.013 & 0.944 ± 0.012 & 0.533 ± 0.035 & 0.598 ± 0.096 \\
            \hline
            MLP & \textcolor{myred}{\xmark} & 0.838 ± 0.032 & 0.917 ± 0.033 & 0.520 ± 0.038 & 0.569 ± 0.088 \\
            MLP & PLR & 0.864 ± 0.015 & 0.954 ± 0.008 & 0.500 ± 0.019 & 0.613 ± 0.055 \\
            MLP & QL & 0.853 ± 0.014 & 0.936 ± 0.011 & 0.520 ± 0.021 & 0.587 ± 0.044 \\
            ResNet & \textcolor{myred}{\xmark} & 0.844 ± 0.034 & 0.926 ± 0.031 & 0.510 ± 0.043 & 0.568 ± 0.058 \\
            ResNet & PLR & 0.866 ± 0.014 & 0.953 ± 0.011 & 0.519 ± 0.032 & 0.564 ± 0.067 \\
            ResNet & QL & 0.865 ± 0.012 & 0.953 ± 0.006 & 0.508 ± 0.027 & 0.593 ± 0.061 \\
            \hline
        \end{tabular}
    }
    \label{tab:sort_week_accuracy}
\end{table*}

\begin{table*}[tb!]
    \centering
    \caption{Comparison of the accuracies for the Sort Prediction task on the day of operations: The column labeled 'All Data' represents the accuracies on the complete testing set, while the remaining columns display accuracies based on a partition of the testing set corresponding to different shift classes of the loads.}
    \renewcommand{\arraystretch}{1.3} 
    \resizebox{0.9\textwidth}{!}{
        \begin{tabular}{c c c c c c}
            \hline
            \multirow{2}{*}{\textbf{Model}} & \multirow{2}{*}{\textbf{Num Emb}} & \multicolumn{4}{c}{\textbf{Sort Prediction Accuracy - Day of Operations}} \\
            \cline{3-6}
             & & \textbf{All Data} & \textbf{No Shift} & \textbf{Internal Shift} & \textbf{External Shift} \\
            \hline
            DT & - & 0.851 ± 0.074 & 0.902 ± 0.083 & 0.648 ± 0.073 & 0.672 ± 0.131 \\
            RF & - & 0.861 ± 0.079 & 0.910 ± 0.086 & 0.671 ± 0.079 & 0.670 ± 0.125 \\
            CatBoost & - & 0.864 ± 0.077 & 0.909 ± 0.081 & 0.679 ± 0.088 & 0.712 ± 0.123 \\
            LGBM & - & 0.922 ± 0.007 & 0.961 ± 0.006 & 0.770 ± 0.027 & 0.776 ± 0.036 \\
            XGBoost & - & 0.925 ± 0.007 & 0.963 ± 0.004 & 0.772 ± 0.021 & 0.783 ± 0.038 \\
            \hline
            MLP & \textcolor{myred}{\xmark} & 0.908 ± 0.022 & 0.951 ± 0.011 & 0.734 ± 0.072 & 0.763 ± 0.057 \\
            MLP & PLR & 0.917 ± 0.009 & 0.957 ± 0.005 & 0.751 ± 0.025 & 0.782 ± 0.053 \\
            MLP & QL & 0.916 ± 0.008 & 0.957 ± 0.004 & 0.755 ± 0.033 & 0.751 ± 0.056 \\
            ResNet & \textcolor{myred}{\xmark} & 0.912 ± 0.033 & 0.958 ± 0.010 & 0.735 ± 0.042 & 0.750 ± 0.071 \\
            ResNet & PLR & 0.922 ± 0.010 & 0.958 ± 0.013 & 0.750 ± 0.039 & 0.764 ± 0.051 \\
            ResNet & QL & 0.916 ± 0.014 & 0.957 ± 0.014 & 0.754 ± 0.035 & 0.781 ± 0.042 \\
            \hline
        \end{tabular}
    }
    \label{tab:sort_day_accuracy}
\end{table*}

\begin{figure*}[t!]
    \centering
    \includegraphics[width=0.99\textwidth]{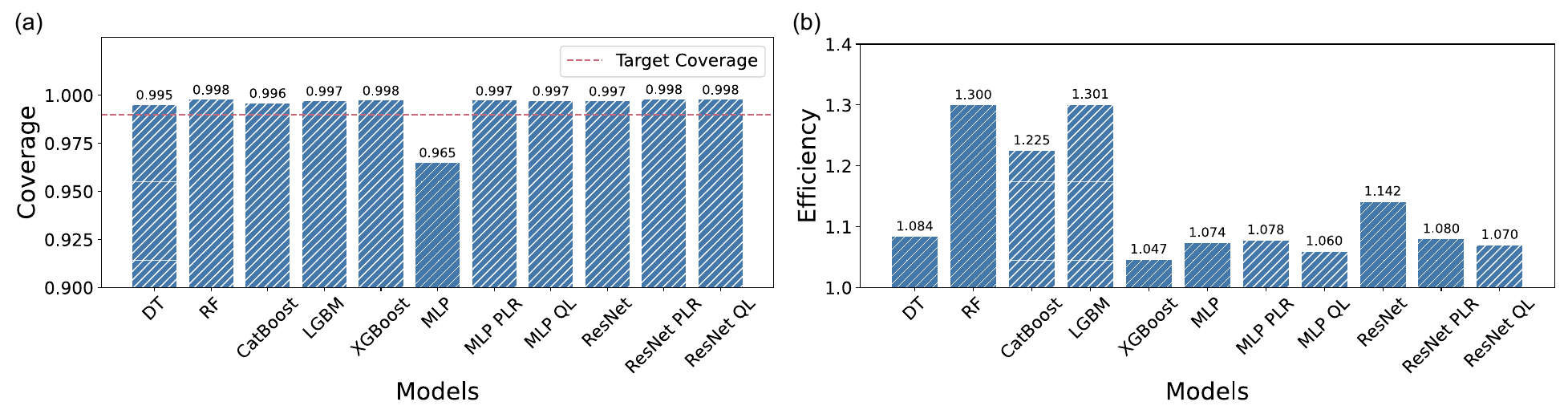}
    \caption{
    Efficiency and coverage per model for the 6-class Building Prediction task: (a) shows the coverage and the fixed target coverage for the building prediction task across the different models. (b) shows the efficiency (average size of the prediction sets) for different models for the building prediction.}
    \label{fig:eff_cov_bldg}
\end{figure*}

\begin{figure*}[t!]
    \centering
    \includegraphics[width=0.99\textwidth]{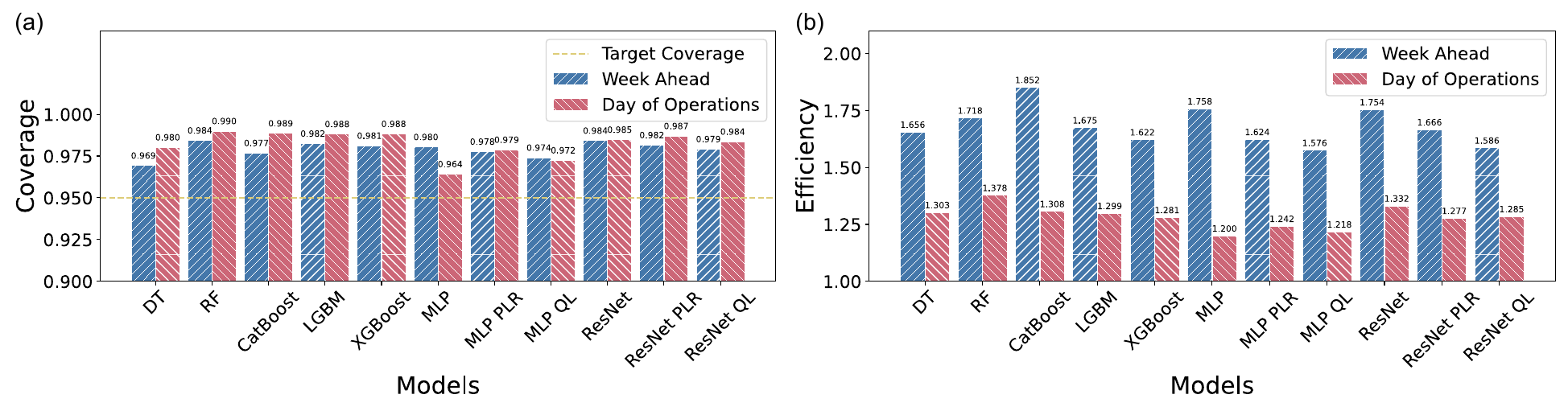}
    \centering
    \caption{Efficiency and coverage per model for the 3-class Sort Prediction task : (a) shows the coverage and the fixed target coverage for the sort prediction task across the different models. (b) shows the efficiency (average size of the prediction sets) for different models for the sort prediction. The blue bar corresponds to the metrics obtained one week before the day of operations and the red one is the one obtained on the day of operations.}
    \label{fig:eff_cov_srt}
\end{figure*}

\subsection{Sort Prediction and Importance of Two-Stage Decision Making}

Table \ref{tab:sort_week_accuracy} illustrates the performance of the
different models on the prediction of the sort, one week before the
day of operations (i.e., stage one of the prediction). While all the
models achieve a global accuracy higher than 80\%, some achieve
significantly higher performance such as ResNet-QL, ResNet-PLR,
MLP-PLR that achieve accuracies exceeding 86\%, slightly outperforming
state of the art traditional models like XGBoost and LightGBM.

As detailed in Section \ref{pbm_statement}, the processing sort can
differ from the planned processing sort in both Internal Shifts and
External Shifts. Table \ref{tab:sort_week_accuracy} reports the
conditional accuracies across the different shift classes. The
accuracy of all models for sort prediction under the No Shift class is
above 90\% (with the exception of DT with 88\% accuracy), which
demonstrates the reliability of the models in predicting the sorts with low
false positive errors in this class. However, the accuracy drops
significantly for the sort prediction in the Internal Shift and
External Shift sub-classes, ranging between 50\% and 60\%. This
reduction is primarily due to the lack of crucial features necessary for
the sort prediction one week before the day of operations.

Even if the initial stage does not yield optimal performance, it is
important for the planners to have an estimate of the network and load
volumes at each building-sort pair one week before the
operations. This early insight helps to identify potential load shifts
and ensures better allocation of personnel and equipment across sorts.

The results for the sort prediction on the day of operation after
incorporating the estimated arrival time of the loads as an additional
feature (i.e., second stage of the decision-making) are presented in
Table \ref{tab:sort_day_accuracy}. A comparison of the results in Table
\ref{tab:sort_week_accuracy} and \ref{tab:sort_day_accuracy} shows
that the inclusion of the additional feature results in a consistent
improvement of approximately $5$-$6\%$ of the overall accuracy across
various models. Moreover, it is noteworthy that the Internal Shift and
External Shift sub-classes exhibit a significant improvement of around
$20\%$ across different models.

\subsection{Conformal Prediction}

As detailed in Section \ref{Method - Conformal Prediction}, conformal prediction is incorporated into each model within the proposed framework. This approach enables the models to generate prediction sets containing multiple buildings or sorts, rather than making a single prediction. This sub-section presents the results in terms of coverage and efficiency for each model, showing that models reach the target coverage, while producing typically small prediction sets. 

The target coverage is set to 99\% (i.e., $\alpha = 0.01$) for the building prediction task due to 1) its significant impact on the tactical planning, and 2) the existing class imbalance in the data which only includes 2\% of the data from the External Shift Class. For the sort prediction, the target coverage is set to 95\% (i.e., $\alpha = 0.05$).

For the building prediction task, the coverage and efficiency of different models are presented in Figure \ref{fig:eff_cov_bldg}. All models reach the target coverage with empirical coverage ranging from 99.5\% to 99.8\%, except for the regular MLP model (with no numerical embedding). This can be attributed to the inherent weakness of the MLP model in this prediction task, resulting in inadequate probability calibration using the calibration data, and insufficient coverage.

Comparing the efficiency of the models on Figure \ref{fig:eff_cov_bldg} reveals that XGBoost achieves the lowest average prediction set size (i.e., efficiency). The proposed DL-based models (with the inclusion of numerical embeddings) also achieve a competitive efficiency ranging between 1.08 to 1.06.

Figures \ref{fig:eff_cov_srt} presents the efficiency and coverage for sort prediction for each model, both one week ahead and on the day of operations. Notably, all models meet the target coverage for sort prediction. Additionally, the models' confidence improves significantly from the week-ahead prediction to the day of operations, as evidenced by the decrease in efficiency (i.e., reduction in the average prediction set size). This highlights the effectiveness of the two-stage decision-making process and the importance of load arrival times in accurately predicting the processing sorts.

\subsubsection{Adaptativeness Property\\}

\begin{figure*}[t!]
    \centering
    \includegraphics[width=0.95\textwidth]{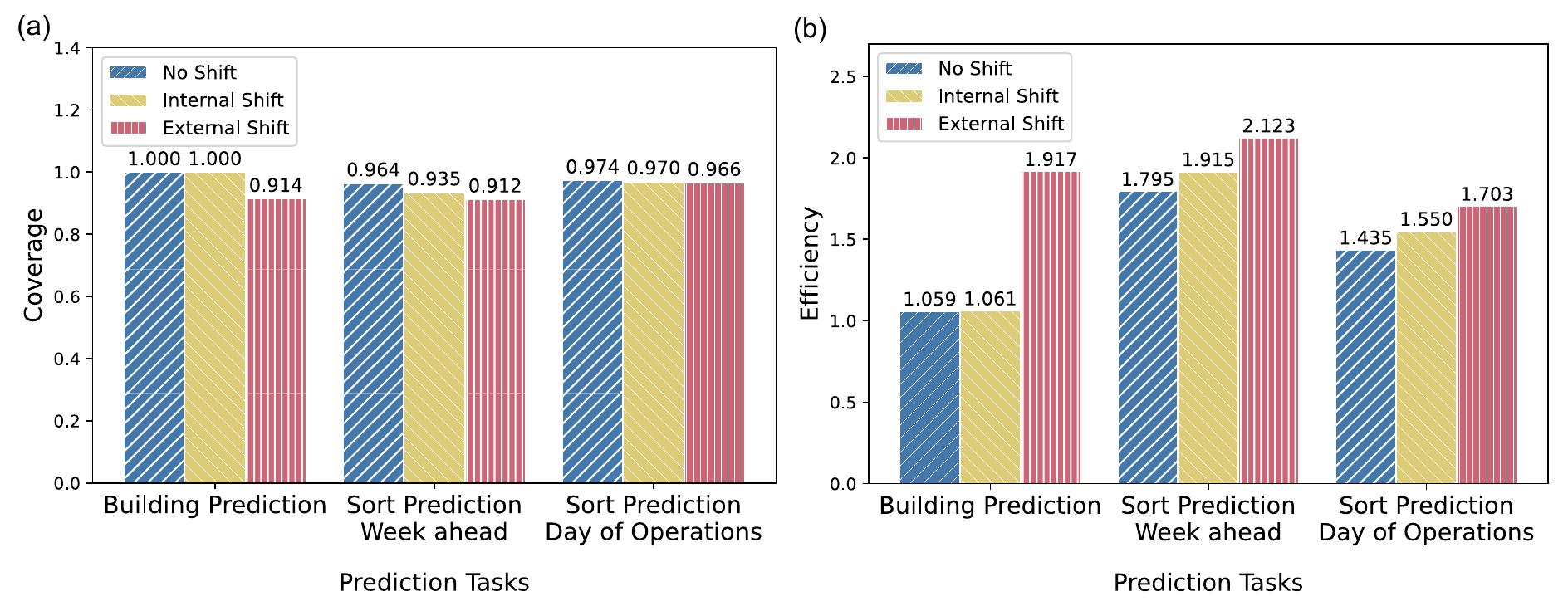}
    \centering
    \caption{Conditional efficiency and coverage for the three different tasks using ResNet-QL: Building Prediction, Sort Prediction - Week ahead, and Sort Prediction - Day of operations. The metrics are illustrated across a partition of the data: No Shift, Internal Shift, and External Shift.}
    \label{fig:adapt_cov_eff}
\end{figure*}

As detailed in Section \ref{Method - Conformal Prediction}, this study
employs RAPS as the scoring method for the uncertainty
quantification. A key feature of RAPS is its adaptive nature, which
adjusts prediction set sizes based on the complexity of the
samples. This adaptability ensures that difficult-to-classify samples
achieve higher efficiency without significantly compromising the
conditional coverage.

Figure \ref{fig:adapt_cov_eff} illustrates the conditional efficiency
across different shift classes for the three tasks under consideration
using the ResNet-QL model (similar results are obtained for the other
methods, however, they are omitted here for simplicity). In the case
of Building Prediction, the conditional efficiency nearly doubles for
samples experiencing external shifts, indicating that the model
effectively captures the complexity of these samples, where the
planned processing building differs from the actual processing
building. For Sort Prediction, conditional efficiency increases by
approximately 7\% between No Shift and Internal Shifts, and by around
11\% from Internal Shifts to External Shifts. This trend underscores
the model ability to accommodate the growing complexity associated
with accurate predictions for shifted loads.

Figure \ref{fig:adapt_cov_eff} presents the corresponding conditional coverages. For Building Prediction, loads from External Shift class are predicted with a conditional coverage of 0.914, demonstrating that the model effectively manages to maintain high coverage, thus supporting accurate shifting decisions. Similar patterns are observed for sort predictions. Notably, the target coverage (set at 0.95) is consistently met across all shift classes for predictions made on the day of operations.

\section{Conclusion and Future Research} \label{sec : Conclusion}

This paper introduced a novel DL-based framework using a two-stage
decision-making to effectively address the inbound load shifting
problem. The experimental results demonstrated the efficacy of the
proposed method in accurately predicting the processing building and
sorts across both majority and minority classes. By leveraging the
theoretical guarantees and enhanced interpretability of conformal
prediction, the proposed framework offers valuable support for
planners in developing effective inbound load plans for both tactical
and operational contexts. Unlike current practices that often rely on
partial network views, the proposed method utilizes historical data to
guide decision making, ensuring more informed and sensible
choices. Additionally, its scalability allows network integration,
enhancing the learning process and improving decision alignment.

The findings highlight the importance of a two-stage decision-making
process for sort prediction. Notably, the results showed an
improvement of around 20\% in accuracy between predictions made one
week before the day of operations and on the day of operations for
Internal and External Shift sub-classes across various
models. Additionally, the second stage of the decision-making process
yielded more confident predictions, highlighting the critical role of
load arrival time in accurately predicting processing sorts.

The proposed DL-based models showed performance comparable to XGBoost,
widely regarded as state of the art on tabular data
analysis. Additionally, the results highlighted the importance and
effectiveness of the embedding layers in enhancing the performance of
the DL-based models. While tuning the parameters of the embedding
layers can introduce challenges during training, the proposed DL
architecture supports transfer learning, allowing for domain
adaptation by using pre-trained models as a warm start. Moreover, the
DL-based model enables the integration of customized loss functions
that account for decision-associated costs.

Future research will focus on integrating the DL model with an
optimization framework to account for decision-associated costs,
including end-to-end training for more optimized decision
making. Additionally, group-conditional conformal prediction will be
explored to enhance the robustness of predictions for minority
classes.

\bibliography{references} 

\newpage

\appendix
\section{Appendix} \label{appendix}

\begin{table}[h!]
\centering
\caption{Tuned Parameters and Ranges for XGBoost and LightGBM}

\begin{tabular}{ll}
\toprule
\textbf{Parameter}     & \textbf{Range}                      \\ \midrule
max depth              & [2, 10]                             \\
learning rate          & [1e-5, 1.0] (log scale)             \\
n estimators           & [50, 500]                           \\
min child weight       & [1, 10]                             \\
alpha                  & [1e-8, 1.0] (log scale)             \\
gamma                  & [1e-8, 1.0] (log scale)             \\
subsample              & [0.5, 1.0] (log scale)              \\
colsample bytree       & [0.01, 1.0] (log scale)             \\
reg alpha              & [1e-8, 1.0] (log scale)             \\
reg lambda             & [1e-8, 1.0] (log scale)             \\ \bottomrule
\end{tabular}
\end{table}

\begin{table}[h!]
\centering
\caption{Tuned Parameters and Ranges for CatBoost}

\begin{tabular}{ll}
\toprule
\textbf{Parameter}      & \textbf{Range}                    \\ \midrule
learning rate           & [1e-5, 1.0] (log scale)           \\
colsample bylevel       & [0.05, 1]                         \\
depth                   & [2, 10]                           \\
min data in leaf        & [1, 100]                          \\ \bottomrule
\end{tabular}
\end{table}

\begin{table}[h!]
\centering
\caption{Tuned Parameters and Ranges for Random Forest}

\begin{tabular}{ll}
\toprule
\textbf{Parameter}       & \textbf{Range}                  \\ \midrule
n estimators             & [50, 500]                       \\
max depth                & [2, 10]                         \\
min samples split        & [2, 30]                         \\
min samples leaf         & [1, 30]                         \\
max features             & [0.001, 1]                      \\ \bottomrule
\end{tabular}
\end{table}

\begin{table}[h!]
\centering
\caption{Tuned Parameters and Ranges for Decision Tree}

\begin{tabular}{ll}
\toprule
\textbf{Parameter}        & \textbf{Range}                \\ \midrule
max depth                 & [1, 100]                     \\
min samples split         & [2, 30]                      \\
min samples leaf          & [1, 30]                      \\ \bottomrule
\end{tabular}
\end{table}

\begin{table}[h!]
\centering
\caption{Tuned Parameters and Ranges for MLP and ResNet}

\begin{tabular}{ll}
\toprule
\textbf{Parameter}   & \textbf{Range}             \\ \midrule
n blocks             & [2, 10]                    \\
d block              & [64, 256]                  \\ \bottomrule
\end{tabular}
\end{table}

\end{document}